\documentclass[conference]{IEEEtran}

\usepackage{times}

\usepackage[numbers]{natbib}
\usepackage{multicol}
\usepackage[bookmarks=true]{hyperref}
\usepackage{xspace}
\usepackage{xcolor}
\usepackage{graphicx}
\usepackage{amsmath,amssymb}
\usepackage{booktabs}
\usepackage{float}
\usepackage{caption}
\usepackage{cleveref}
\usepackage{bm}
\usepackage{graphicx}

\crefname{figure}{Fig.}{Figs.}

\crefname{table}{Table}{Tables}
\crefname{section}{Sec.}{Secs.}
\crefname{equation}{Eq.}{Eqs.}

\renewcommand{\baselinestretch}{0.99}

\newcommand{\method}{\textsc{PHP}\xspace}

\begin{document}

\title{Perceptive Humanoid Parkour: Chaining \\ Dynamic Human Skills via Motion Matching}

\author{Zhen Wu*$^{1}$, Xiaoyu Huang*$^{1, 2}$, Lujie Yang*$^{1}$, Yuanhang Zhang$^{1, 3}$, Xi Chen$^{ 1}$, \\ Pieter Abbeel$^{\dagger 1, 2}$, Rocky Duan$^{\dagger 1}$, Angjoo Kanazawa$^{\dagger 1, 2}$, Carmelo Sferrazza$^{\dagger 1}$, Guanya Shi$^{\dagger 1, 3}$, C. Karen Liu$^{\dagger 1, 4}$ \\
$^{1}$Amazon FAR,
$^{2}$UC Berkeley, 
$^{3}$CMU,
$^{4}$Stanford University,
$^\dagger$Amazon FAR team co-lead\\
Page: \href{https://php-parkour.github.io/}{\texttt{\color{magenta}https://php-parkour.github.io/}}%
}



%


\twocolumn[{%
\renewcommand\twocolumn[1][]{#1}%
\maketitle

    \includegraphics[width=\textwidth]{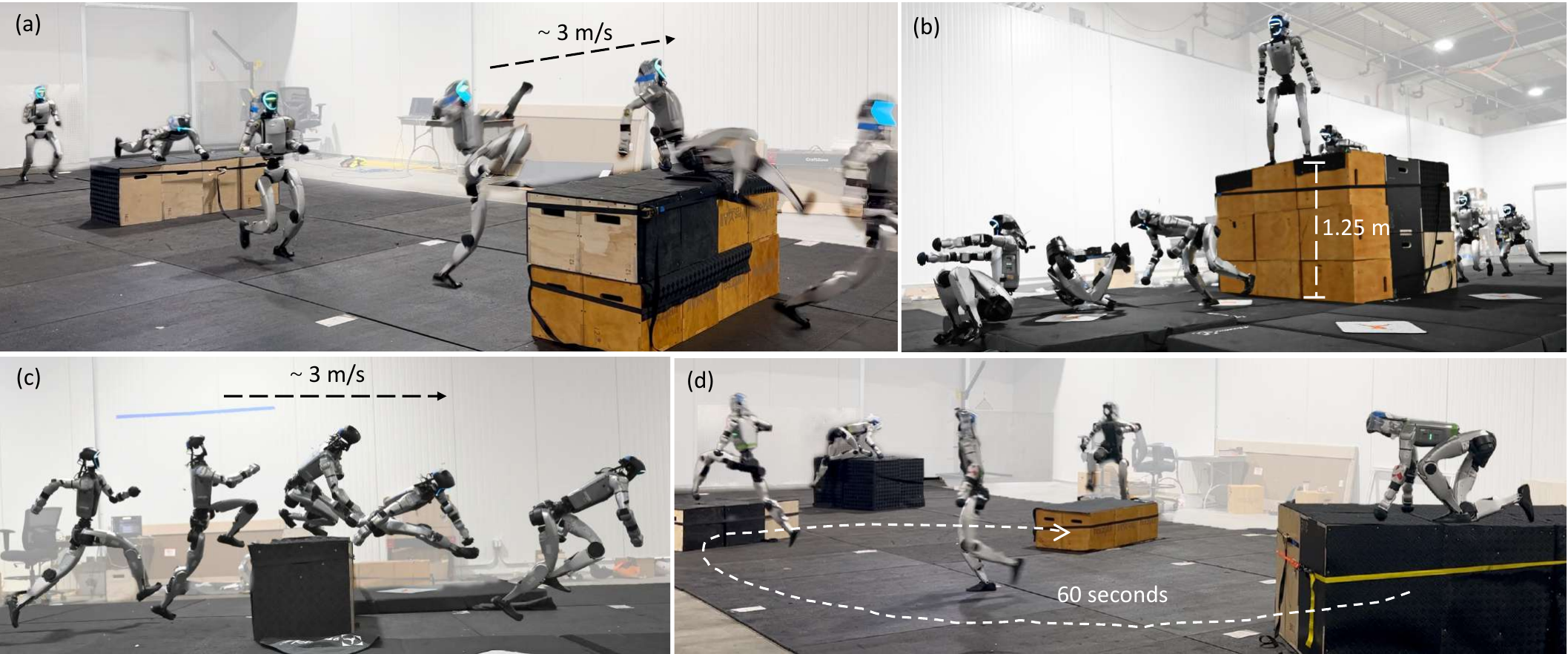}
    \captionof{figure}{\textbf{Perceptive Humanoid Parkour (\method)} enables a Unitree G1 humanoid robot to execute highly dynamic, long-horizon parkour behaviors using onboard perception. By composing various agile human skills via motion matching and a teacher-student training pipeline, we train a single multi-skill visuomotor policy capable of complex contact-rich maneuvers including (a) cat-vaulting over a short obstacle followed by dash-vaulting over a higher obstacle at approximately 3\,m/s, (b) climbing onto a 1.25\,m (96\% of robot height) wall, and rolling down, (c) speed-vaulting over an obstacle at approximately 3\,m/s, and (d) a 60-second continuous traversal of a complex parkour course with autonomous skill selection and seamless transitions.}
    \label{figure1}
}]

\begingroup
  \renewcommand\thefootnote{}
  \footnotetext{* Equal contribution.}
\endgroup

\begin{abstract}
While recent advances in humanoid locomotion have achieved stable walking on varied terrains, capturing the agility and adaptivity of highly dynamic human motions remains an open challenge. In particular, agile parkour in complex environments demands not only low-level robustness, but also human-like motion expressiveness, long-horizon skill composition, and perception-driven decision-making. In this paper, we present Perceptive Humanoid Parkour (\method), a modular framework that enables humanoid robots to autonomously perform long-horizon, vision-based parkour across challenging obstacle courses. Our approach first leverages motion matching, formulated as nearest-neighbor search in a feature space, to compose retargeted atomic human skills into long-horizon kinematic trajectories. This framework enables the flexible composition and smooth transition of complex skill chains while preserving the elegance and fluidity of dynamic human motions. Next, we train motion-tracking reinforcement learning (RL) expert policies for these composed motions, and distill them into a single depth-based, multi-skill student policy, using a combination of DAgger and RL. Crucially, the combination of perception and skill composition enables autonomous, context-aware decision-making: using only onboard depth sensing and a discrete 2D velocity command, the robot selects and executes whether to step over, climb onto, vault or roll off obstacles of varying geometries and heights. We validate our framework with extensive real-world experiments on a Unitree G1 humanoid robot, demonstrating highly dynamic parkour skills such as climbing tall obstacles up to 1.25\,m (96\% robot height), as well as long-horizon multi-obstacle traversal with closed-loop adaptation to real-time obstacle perturbations.

\end{abstract}

\IEEEpeerreviewmaketitle

\section{Introduction}
Achieving the agility and adaptivity of human motion in traversing complex terrains remains a central challenge for humanoid robotics. Humans traverse challenging terrains of drastically different dimensions by rapidly selecting and chaining dynamic whole-body skills based on perceived environmental context. Our goal is to endow humanoids with the same capability. In this work, we study parkour as a concrete, self-contained testbed for this broader objective.

Parkour highlights several core challenges. First, the robot must perform highly dynamic and contact-rich skills, such as climbing walls around or above its body height or vaulting over obstacles within fractions of a second. This requires effective control in the humanoid's vast, high-dimensional action space. Second, these skills must be tightly coupled with exteroception, such as vision, to enable adaptation to environmental variation and rapid reaction to unexpected perturbations. Furthermore, to generalize beyond isolated maneuvers and traverse complex obstacle courses, 
the robot must consolidate many highly dynamic skills into a single visuomotor policy, which becomes increasingly difficult as the number and diversity of required skills grow.  

Human motion data has become essential for learning highly dynamic humanoid behaviors. Prior work~\cite{liao2025beyondmimic, yang2025omniretarget} has used human motion data to successfully demonstrate highly dynamic skills such as jumping, rolling, and flipping.
However, highly dynamic motion data is inherently scarce: capturing fast, contact-rich maneuvers typically requires specialized setups and careful curation, so datasets often include only one or two demonstrations per skill, each lasting just a few seconds. This scarcity is not unique to parkour but applies broadly to all dynamic human skills. Yet long-horizon tasks such as parkour require both rich within-skill variation that adapts to how the robot approaches an obstacle, and smooth, natural transitions between multiple skills across complex courses.

To address this challenge, we adopt motion matching \cite{Buettner2015MotionMatchingVideo, Clavet2016MotionMatching} 
as a simple yet powerful mechanism.
Motion matching synthesizes long-horizon motion by retrieving and stitching motion fragments via nearest-neighbor search in a designed feature space. Crucially, this process densifies a sparse motion library by producing diverse transitions across approach distances, headings, and timing, while preserving the realism of captured motions. In our framework, motion matching enables the generation of a large set of obstacle-adaptive, long-horizon kinematic reference trajectories for downstream policy learning. \looseness=-1

Learning a visuomotor policy that executes dozens of highly dynamic skills requires perceptive inputs that can be efficiently simulated and reliably transferred to the real world. 
To improve training efficiency, prior work typically trains privileged state-based experts in simulation and distills them into vision-based students using DAgger~\cite{ross2011reduction}. However, for humanoid parkour, pure imitation loss is limited: compounding errors can quickly derail highly dynamic skills such as climbing and vaulting. To address this, we augment distillation with an RL objective that provides task-level corrective feedback, steering the student towards successful traversal and yielding a scalable recipe across many skills.

To this end, we present Perceptive Humanoid Parkour (\method), a modular framework that integrates human motion priors, long-horizon skill composition, and perceptive control. We first retarget human motion data into a library of robot-compatible atomic skills using OmniRetarget \cite{yang2025omniretarget}. We then employ motion matching to compose these skills into a diverse set of long-horizon kinematic trajectories. These composed trajectories preserve agility and smooth transitions, while providing sufficient variation to learn adaptive long-horizon behaviors. We then train motion-tracking expert policies and distill them into a single depth-conditioned, multi-skill policy that enables the robot to autonomously select and transition among behaviors, such as stepping, climbing, and vaulting, using onboard depth sensing.

Our contributions are threefold:
\begin{enumerate}
    \item An efficient kinematic skill composition pipeline that chains retargeted human motions into diverse long-horizon trajectories via motion matching.
    \item A scalable training framework that distills multiple experts into a single visuomotor policy, enabling seamless transitions across diverse parkour skills.
    \item Successful zero-shot sim-to-real transfer of depth-based policies on a physical humanoid robot, achieving highly dynamic parkour over various obstacles.
\end{enumerate}
\section{Related Works}

The goal of parkour is to traverse challenging terrains agilely by perceiving, reacting, and chaining skills for different obstacles. We review related work in these areas.

\subsection{Perceptive Terrain Traversal for Legged Robots}
While blind locomotion has achieved strong robustness on moderately structured terrains such as slopes and stairs on quadrupedal robots~\cite{lee2020learning, nahrendra2023dreamwaq, long2023hybrid, kumar2021rma}, perception enables traversal of substantially more challenging terrains~\cite{miki2022learning}. 
In particular, perception is critical for handling sparse footholds~\cite{agarwal2023legged, yang2023neural, yu2024walking} and discontinuous terrain such as gaps and tall obstacles~\cite{agarwal2023legged, yu2024learning}. Building on these capabilities, prior work has enabled quadrupeds to traverse parkour-style terrain courses with consecutive gap jumps and obstacle climbs~\cite{cheng2024extreme, luo2024pie, hoeller2024anymal, rudin2025parkour}. 

However, translating the success on quadrupeds to humanoids remains challenging. While quadrupedal parkour skills can often be trained from scratch via reward shaping, this approach scales poorly to humanoids due to high-dimensional whole-body control. As a result, prior perceptive humanoid locomotion has primarily focused on lower-dynamic terrain traversal, including stair climbing~\cite{long2025learning, yang2026locomotion}, walking on sparse terrain~\cite{wang2025beamdojo, he2025attention, ben2025gallantvoxelgridbasedhumanoid}, and stepping onto low platforms~\cite{zhuang2024humanoid}. 
Moreover, 
to reduce exploration difficulty in RL when training from scratch, most works adopt a teacher-student pipeline where an expert is trained with privileged states and a vision-based student is distilled via DAgger~\cite{cheng2024extreme, rudin2025parkour}. We follow this paradigm but find pure DAgger insufficient for highly dynamic humanoid skills, and therefore augment it with RL to improve distillation performance. 
Note that this differs from the fine-tuning stage in~\cite{rudin2025parkour}, which primarily focuses on adapting an already performant DAgger-distilled policy to unseen terrains.

\begin{figure*}
    \centering
    \includegraphics[width=\linewidth]{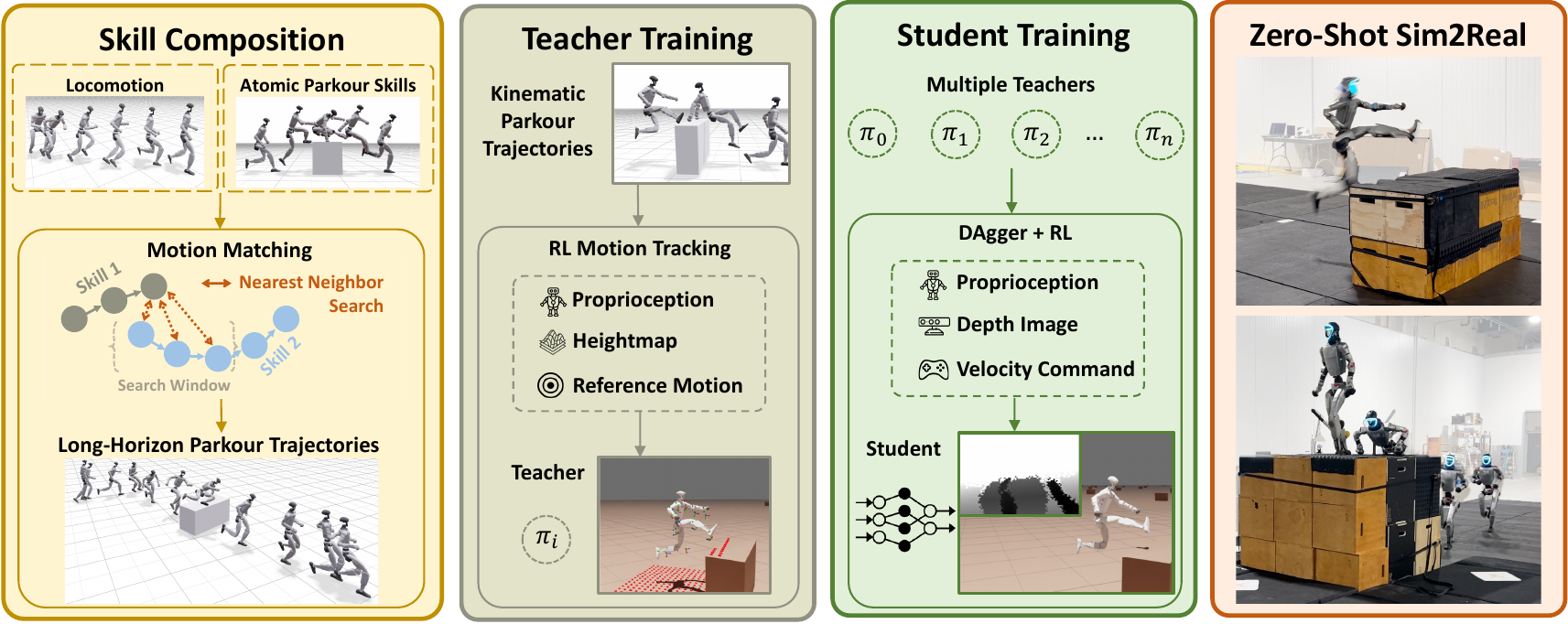}
    \caption{\textbf{Perceptive Humanoid Parkour overview.} Atomic parkour skills are composed into long-horizon kinematic reference trajectories via motion matching. Single-skill teacher policies are trained with privileged information using RL-based motion tracking. Multiple teachers are distilled into a single depth-based student policy using a hybrid DAgger and RL objective. This scalable recipe enables zero-shot sim-to-real transfer onto a physical humanoid robot that adaptively traverses through complex terrains by autonomously executing highly agile parkour skills using onboard perception. 
    }
    \label{fig:overview}
    \vspace{-10pt}
\end{figure*}

\subsection{Humanoid Skill Chaining with Human Motion Data}


Using human motion references effectively reduces reward engineering and produces agile, natural humanoid behaviors~\cite{liao2025beyondmimic, yang2025omniretarget, zhang2025hub, pan2025agility, xie2025kungfubot, chen2025gmt}, but comes at the cost of more challenging skill chaining. With reward shaping, quadrupeds can learn transitions either implicitly by a single policy~\cite{cheng2024extreme, luo2024pie, he2025attention, ben2025gallantvoxelgridbasedhumanoid}, or through specialist switching or distillation using a shared locomotion state~\cite{hoeller2024anymal, zhuang2023robot, caluwaerts2023barkour, rudin2025parkour}. In contrast, human motion data spans heterogeneous styles that can lie in disjoint regions of the state space, making long-horizon composition a fundamental challenge.

AMP~\cite{peng2021amp} addresses this challenge by training a single policy to learn a distribution of skills, allowing transitions to implicitly emerge from RL exploration, but replaces hand-crafted rewards with a learned style reward from motion data. While promising in animation and quadrupeds~\cite{xu2025learning, wu2023learning}, humanoid hardware demonstrations have so far been limited to less agile skills such as walking, stepping, and box lifting~\cite{zhu2026hiking, sun2025dpl, wang2025physhsi}.




To address the transition problem more explicitly, another line of work generates intermediate kinematic trajectories using learned kinematics models (e.g., MDM~\cite{tevet2023human}) and executes them with tracking controllers (e.g., DeepMimic~\cite{peng2018deepmimic}). These kinematics models can provide smooth transition references at test time~\cite{xu2025parc, luo2025sonic, fu2024humanplus, ze2025twist2} or training time~\cite{kalaria2025dreamcontrol}, but their trajectory quality degrades significantly in the low-data regimes common in parkour. This often requires either costly iterative co-training~\cite{xu2025parc} to recover usable motion or receding-horizon replanning~\cite{huang2025diffuse}, which is costly with perception in real time.\looseness=-1

In contrast, we adopt motion matching~\cite{Buettner2015MotionMatchingVideo, holden2020mm} as a simple yet highly effective source of kinematic references for humanoid skill chaining. 
Motion matching has been widely adopted in video games and character animation for its simplicity and practical controllability, while still producing high-quality motion~\cite{Buettner2015MotionMatchingVideo, gou2025control}.
While a mature technique in animation~\cite{bergamin2019drecon}, it has so far been applied in robotics only to relatively simple quadruped behaviors~\cite{kang2021animal}. In this work, we show that it is a powerful tool for chaining dynamic and expressive human skills over difficult terrain courses for humanoid robots, substantially improving both success rate and transition smoothness.







\section{Adaptive and Agile Long-Horizon Parkour}

\subsection{Overview}
\label{subsec:overview}
The objective of this work is to enable a humanoid robot to execute agile parkour behaviors over multiple obstacles autonomously using onboard perception. We first generate long-horizon kinematic reference motions via motion matching by composing locomotion with atomic parkour skills.
We then train motion-tracking expert policies with privileged observations in simulation, and finally distill them into a depth-based student policy using DAgger in combination with a PPO objective, enabling zero-shot sim-to-real deployment.
An overview of the system is shown in \cref{fig:overview}.

\subsection{Skill Composition via Motion Matching}

Motion matching \cite{Buettner2015MotionMatchingVideo,Clavet2016MotionMatching} is a technique originally developed in the video game industry for interactive character control, where motion is generated online by selecting, at each frame or transition point, the animation frame from a large database whose motion features best match the current pose and desired future behavior.
In this work, we adopt motion matching as an offline motion synthesis module for composing scarce atomic parkour skills with locomotion into long-horizon references.  

\subsubsection{Basic motion matching}\label{sec:basic_mm}
We briefly summarize the standard motion matching formulation;  implementation details are provided in Appx.~\ref{appx:mm_implementation}.
Let a motion database consist of $N$ frames, where each frame $i$ is associated with a kinematic pose $\bm{q}_i$ and a matching feature vector $\bm{x}_i$ derived from $\bm{q}_i$.
 Following~\cite{holden2020mm}, $\bm{x}_i$ concatenates (i) short-horizon future trajectory positions and facing directions, (ii) local foot joint positions and velocities, and (iii) root velocity, all expressed in the character's local coordinate frame.

When generating transitions, given the current character state and a desired 2D velocity command, we first convert the command into a desired future trajectory and facing directions, and then concatenate these with foot positions, velocities and root velocities from the current state to form the query feature $\hat{\bm{x}}_t$.
The best matching frame is then retrieved via
\begin{equation}
i_t^\star = \arg\min_{i \in \mathcal{C}_t} \; \| \hat{\bm{x}}_t - \bm{x}_i \|^{2},
\label{eq:mm_nn}
\end{equation}
where $\mathcal{C}_t$ denotes the search window of the user-specified upcoming skill. 
This nearest-neighbor search is performed periodically (every $M$ frames) or when the commanded velocity changes significantly. 
After selecting $i_t^\star$, the system transitions playback to frame $i_t^\star$ and plays forward from that frame until the next search is performed.
A short blending window is applied around transitions to avoid discontinuities.


\subsubsection{Long-Horizon Parkour Trajectory Synthesis}
Since locomotion (walking and running) serves as a ubiquitous and naturally reusable connector between more challenging parkour skills, we generate long-horizon parkour trajectories by composing locomotion segments with short parkour skill clips in the form of
$\texttt{Locomotion}$ $\rightarrow$ $\texttt{Parkour Skill}$ $\rightarrow$ \texttt{Locomotion}.
By routing all skills through a shared locomotion manifold, this formulation enables consistent transitions across heterogeneous skills without requiring specific, hand-captured transitions between every possible skill pair, and supports scalable composition of long-horizon behaviors.

We maintain (i) a locomotion database $\mathcal{D}_{\text{loco}} = \{(\bm{x}_i^{\text{loco}}, \bm{q}_i^{\text{loco}})\}$ and (ii) a set of skill databases $\{\mathcal{D}_{k}\}$, one for each parkour skill $k$.
Each skill clip is paired with a corresponding terrain asset.
For every atomic skill clip in $\mathcal{D}_k$, we manually annotate the skill start and end frame indices $(s_k, e_k)$.
We additionally define a pre-skill entry window of skill-dependent length $H_{k}$:
\begin{equation}
\mathcal{E}_k := [\, s_k - H_k,\; s_k \,],
\label{eq:entry_window}
\end{equation}
which corresponds to the approach phase right before the main contact-rich maneuver, where transitioning into the clip is meaningful. 
For example, for a vault clip, $\mathcal{E}_k$ captures the final approach steps before takeoff, avoiding transitions outside the intended approach phase.



\textbf{Locomotion mode.}
During locomotion, we run standard motion matching via Eq.\eqref{eq:mm_nn} with $\mathcal{C}_t = \mathcal{D}_{\text{loco}}$, and advance playback sequentially as described in Sec.~\ref{sec:basic_mm}.


\textbf{Locomotion $\rightarrow$ Skill transition.}
When a transition into skill $k$ is required, we restrict the search window $\mathcal{C}_t$ to the pre-skill entry window $\mathcal{E}_k$, and transition to the matched entry frame through Eq.\eqref{eq:mm_nn}. 
After the transition, the skill clip is replayed sequentially until the annotated end frame $e_k$.
At the switch, we place the paired terrain by applying the terrain-to-root offset at the matched entry frame in the reference clip to the robot’s current root pose.
During skill execution, we disable further motion matching and simply advance the playback index to preserve the contact-rich human motion.

\textbf{Skill $\rightarrow$ Locomotion transition.}
After reaching $e_k$, we return to locomotion by resuming motion matching via Eq.\eqref{eq:mm_nn} with $\mathcal{C}_t = \mathcal{D}_{\text{loco}}$, and continue sequential playback.

\begin{figure}
    \centering
    \includegraphics[width=\linewidth]{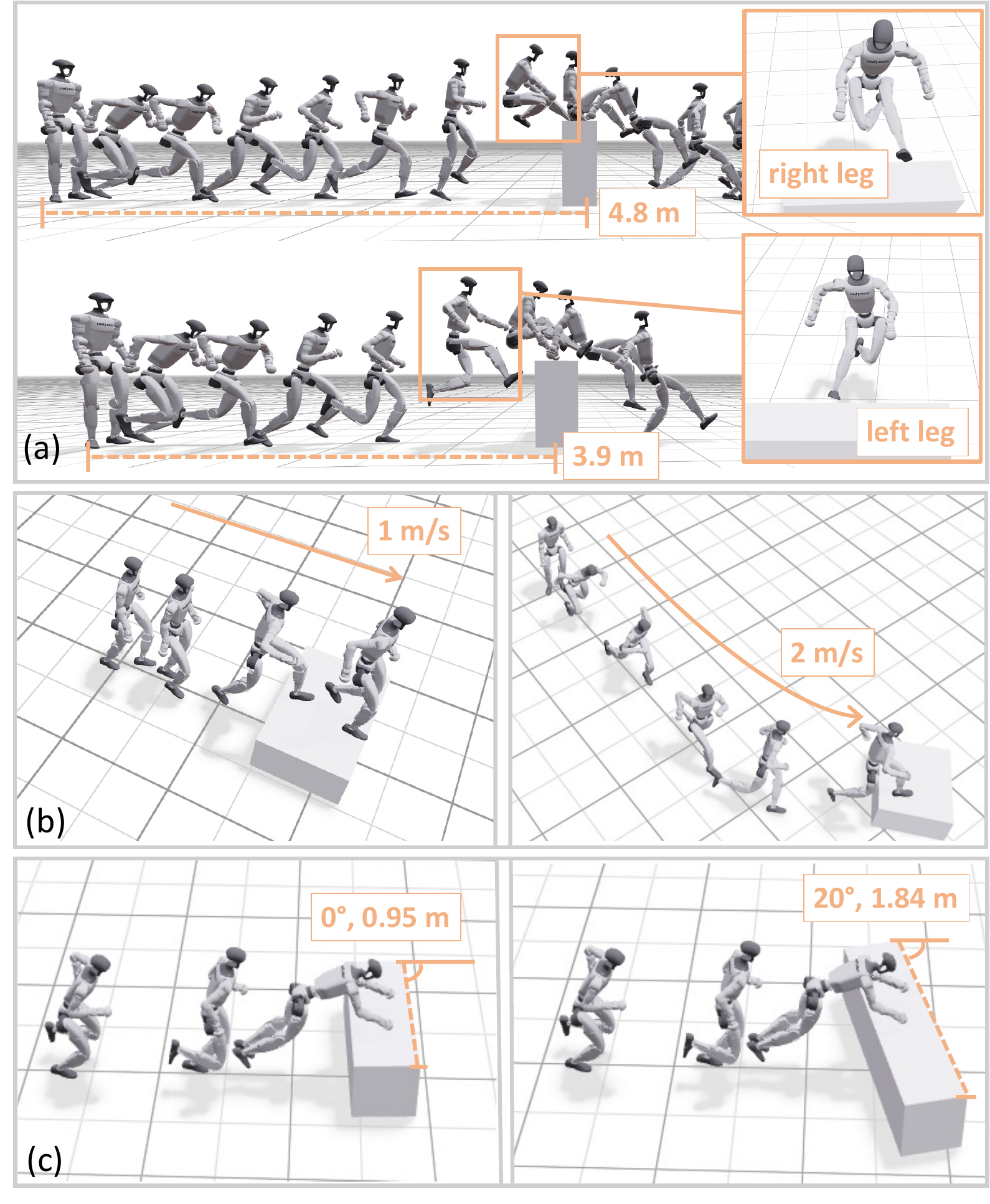}
    \vspace{-0.5cm}
    \caption{\textbf{Diverse variations of composed parkour skills synthesized via motion matching. }
    (a) Different approach distances trigger varying stride phases and entry poses. 
    (b) Diverse locomotion speeds, directions, and durations. (c) Randomized terrain poses and shapes. 
    }
    \label{fig:randomization}
    \vspace{-0.5cm}
\end{figure}

\textbf{Dataset construction.}
\label{sec:dataset_construction}
We synthesize long-horizon reference trajectories by rolling out the motion-matching composition procedure as follows.
As visualized in \cref{fig:randomization}(b), each trajectory starts from a standing state and enters a locomotion phase driven by 2D velocity commands sampled from two speed levels (\texttt{low} (1m/s), \texttt{high} (2m/s)) 
and five turning directions
($-90^\circ$, $-45^\circ$, $0^\circ$, $45^\circ$, $90^\circ$).
We then transition into skill $k$ and replay the skill clip sequentially; during the skill, we set the command to go straight ($0^\circ$) while keeping the same speed level as the preceding locomotion segment.
After reaching the annotated end frame $e_k$, we return to locomotion and continue for an additional $2$ seconds before stopping.
Throughout synthesis, we record the per-timestep velocity commands alongside the generated kinematic reference poses $\{\bm{q}_t\}$, and use these paired trajectories for subsequent policy training. \looseness=-1

\textbf{Transition Density.} Motion matching naturally induces a high density of transitions by allowing a skill to be entered from multiple locomotion states that are nearby in the motion feature space. We exploit this to generate diverse skill entrances spanning different approach distances and stride phases (e.g., adding a preparatory step before a jump, or initiating a vault from different phases of a running gait), densifying the distribution of pre-skill states. As illustrated in  \cref{fig:randomization} (a), varying the initial approach distance (e.g., $3.9$ m vs. $4.8$ m) forces the motion matching engine to select different stride sequences, resulting in distinct entry poses such as left-leg versus right-leg leads. To prevent non-causal shortcuts (e.g., relying on elapsed time or step count), we randomize the pre-skill locomotion duration by sampling it uniformly from $[0.1, 3]$\,s, with an average interval of $0.3$\,s. Such diverse motion-terrain pairs encourage context-based reaction, and are critical for learning a policy that can reliably trigger the correct skill under varying distances and timings.

\textbf{Terrain Randomization.} 
To improve robustness beyond the training obstacles while keeping the reference feasible, we randomize obstacle geometry and pose around each synthesized trajectory. Specifically, obstacle width is sampled from the minimum required by the reference up to 1.5 m; the remaining dimensions are perturbed within $\pm 5$ cm; and obstacle yaw is randomized within $\pm 45^\circ$, as illustrated in Fig. \ref{fig:randomization}(c). This exposes the policy to variations in obstacle shape and pose without invalidating the underlying reference.

\textbf{Distractors.} We place distractor boxes with random sizes and poses near the reference trajectory to improve robustness to irrelevant objects and reduce overfitting in the real world. 

\subsection{Learning a Highly-Dynamic Visuomotor Policy}
Our goal is to train a single perceptive policy capable of various long-horizon parkour skills. Commanded by a target velocity, the humanoid will autonomously perform various parkour skills based on the obstacles it perceives.
Because the skills are highly dynamic, we train skill-specific experts to achieve high motion quality and then distill them into a single visuomotor policy. To ensure scalability, we use a unified expert and distillation formulation without motion-specific tuning. \looseness=-1

\subsubsection{Training Expert Policies with Motion Tracking}
We follow BeyondMimic~\cite{liao2025beyondmimic} and OmniRetarget~\cite{yang2025omniretarget} for motion tracking, and refer readers to these prior works for details.

\textbf{Observations} include reference joint position/velocity, reference pelvis pose error, pelvis linear/angular velocity, joint position/velocity, and the previous action. We additionally provide the expert with a 0.7 m × 0.7 m height scan, allowing it to adapt to terrain randomizations.

Unlike~\cite{yang2025omniretarget}, we enable global tracking with privileged observations (pelvis global position and velocity) so the expert can learn recovery behaviors. This is important because the reference motion is tightly coupled with the terrain, meaning small drift or timing errors can quickly accumulate and must be corrected to stay on the intended trajectory. While these privileged states are not available on hardware, they can be inferred from visual inputs by the student policy.

\textbf{Adaptive Sampling} is essential for learning difficult skills in expert training, which prioritizes sampling from regions that fail more frequently.  
For example, without it, the high-wall climbing expert fails to converge to a meaningful behavior.

\textbf{Rewards, Terminations, and Domain Randomization} follow BeyondMimic~\cite{liao2025beyondmimic}: DeepMimic-style tracking rewards with action rate, joint limits, and collision penalties, tracking-based early termination, and lightweight randomizations.

\textbf{Actions} are joint PD targets normalized by a fixed action scale. Due to challenging RL exploration, we set the action scale to 1 for all experts, instead of the heuristics used in~\cite{liao2025beyondmimic}.

\subsubsection{Distilling a Unified Student Policy with DAgger and RL}
\label{sec:distillation_method}
A common approach for learning a unified policy from multiple experts is to apply DAgger-style imitation learning~\cite{ross2011reduction, cheng2024extreme, zhuang2023robot, rudin2025parkour, liao2025beyondmimic}. While effective for easier motions such as stepping, we find that DAgger alone is insufficient for highly dynamic skills such as climbing and vaulting. 
These skills depend on brief, high-magnitude torque bursts, but per-step imitation objectives like DAgger do not account for episode outcomes and therefore do not explicitly favor such high-torque actions.
For example, actions that result in higher or lower root positions that are symmetric about the reference may receive identical DAgger loss, even though only the higher-root trajectory successfully clears the obstacle.

To address this, we apply PPO alongside DAgger with a curriculum,
\begin{equation}
\mathcal{L}=\lambda_{\text{PPO}}\,\mathcal{L}_{\text{PPO}}+\lambda_{D}\,\mathcal{L}_{D},
\qquad \lambda_{\text{PPO}}+\lambda_{D}=1,
\label{eq:rl_dagger}
\end{equation}
where $\lambda_{\text{PPO}}$ and $\lambda_{D}$ are their curriculum weights.
Note that the primary role of PPO is to provide a success-driven signal that encourages exploiting expert behaviors, such as high-torque actions,
rather than exploring beyond the expert skill distribution. This hybrid setup substantially improves the unified policy’s performance on diverse, highly dynamic skills.

\textbf{Observations, Actions, and Domain Randomization.}
The policy observes proprioception signals including pelvis gravity vector and angular velocity, joint positions and velocities, and the previous action. For vision, we use depth images rendered with Nvidia WARP~\cite{warp2022} for high-throughput training. The policy also receives velocity commands defined in Sec.~\ref{sec:dataset_construction}. The action space and domain randomization for sim-to-real transfer are identical to those used in expert training setup. 

\textbf{Camera modeling and depth artifacts.} We calibrate the simulated camera by matching robot self-visibility across a set of poses in simulation and hardware using ROI overlap, and randomize camera extrinsics within 2.5 cm translation and 2.5° rotation around the calibrated value to improve robustness to viewpoint shifts and mounting variability. We inject realistic depth noise following prior work~\cite{rudin2025parkour}, but exclude Gaussian blur since it can obscure obstacles at high speed. Finally, we randomize observation delay between 60 ms and 80 ms to simulate hardware latency fluctuations.

\textbf{Curriculum. }
Since PPO gradients are noisy in the early stages and can otherwise undermine distillation, we apply a warmup curriculum that gradually shifts from DAgger to PPO. The curriculum includes three parts.

First, we linearly tune down $\lambda_D$ in \cref{eq:rl_dagger} during the first half of training, capped at 0.1: $\lambda_D(k)=\max\left(0.1,1-\frac{k}{K/2}\right)$, where $k$ is the current iteration and $K$ is the total iterations. \looseness=-1

Second, left-right symmetry introduces multimodality: many skills admit two equally valid mirrored executions, for example, clearing a hurdle with either the left- or right-leg lead, while the reference trajectory represents only one of them. As a result, the distilled policy may perform the mirrored mode, which still completes the skill but incurs a large tracking error and would be terminated incorrectly. This spurious failure signal can cause high reward variance for PPO. To mitigate this issue, we relax the termination threshold from 0.5\,m for the expert to 1\,m using the same linear schedule, so mirrored modes are not terminated prematurely.

Finally, we enable adaptive learning rate and KL-based exploration control only when $\lambda_{\text{PPO}}$ exceeds 0.1.

\textbf{Adaptive Sampling} of rollout start points is disabled during student training. While it helps experts focus on failures to learn more difficult segments, it can undersample ``borderline'' clips that do not fail in simulation but exhibit jittery behavior, which often leads to large sim-to-real gap on hardware. To further avoid data imbalance across skills, we sample each skill evenly and sample uniformly within each skill.
\section{Experiments}
We evaluate the proposed framework through a series of simulation and real-world experiments on a Unitree G1 humanoid (1.3\,m tall with 29 DoFs). For training, we use a 3-layer CNN and a 5-layer MLP with hidden sizes [2048, 1024, 512, 256, 128], trained with 16,384 parallel environments. Both expert and student policies are trained for 20K iterations. 

\begin{figure}
    \centering
    \includegraphics[width=\linewidth]{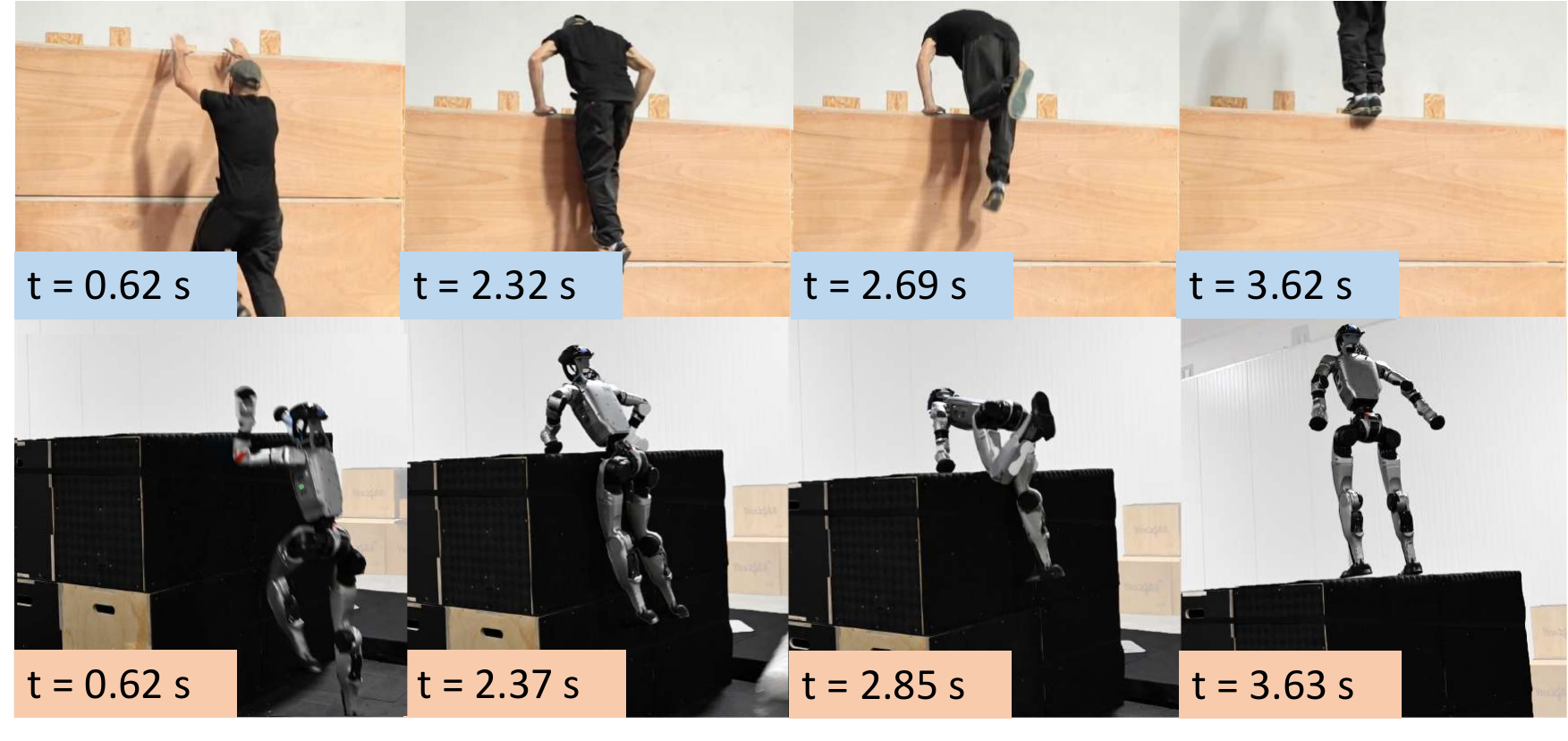}
    \vspace{-0.6cm}
    \caption{Side-by-side comparison of high-climb agility. The robot climbs onto a 1.25\,m wall within 3.63\,s.
    }
    \label{fig:climb_134}
    \vspace{-0.5cm}
\end{figure}
\subsection{Real-World Results}

\begin{figure*}
    \centering
    \includegraphics[width=\linewidth]{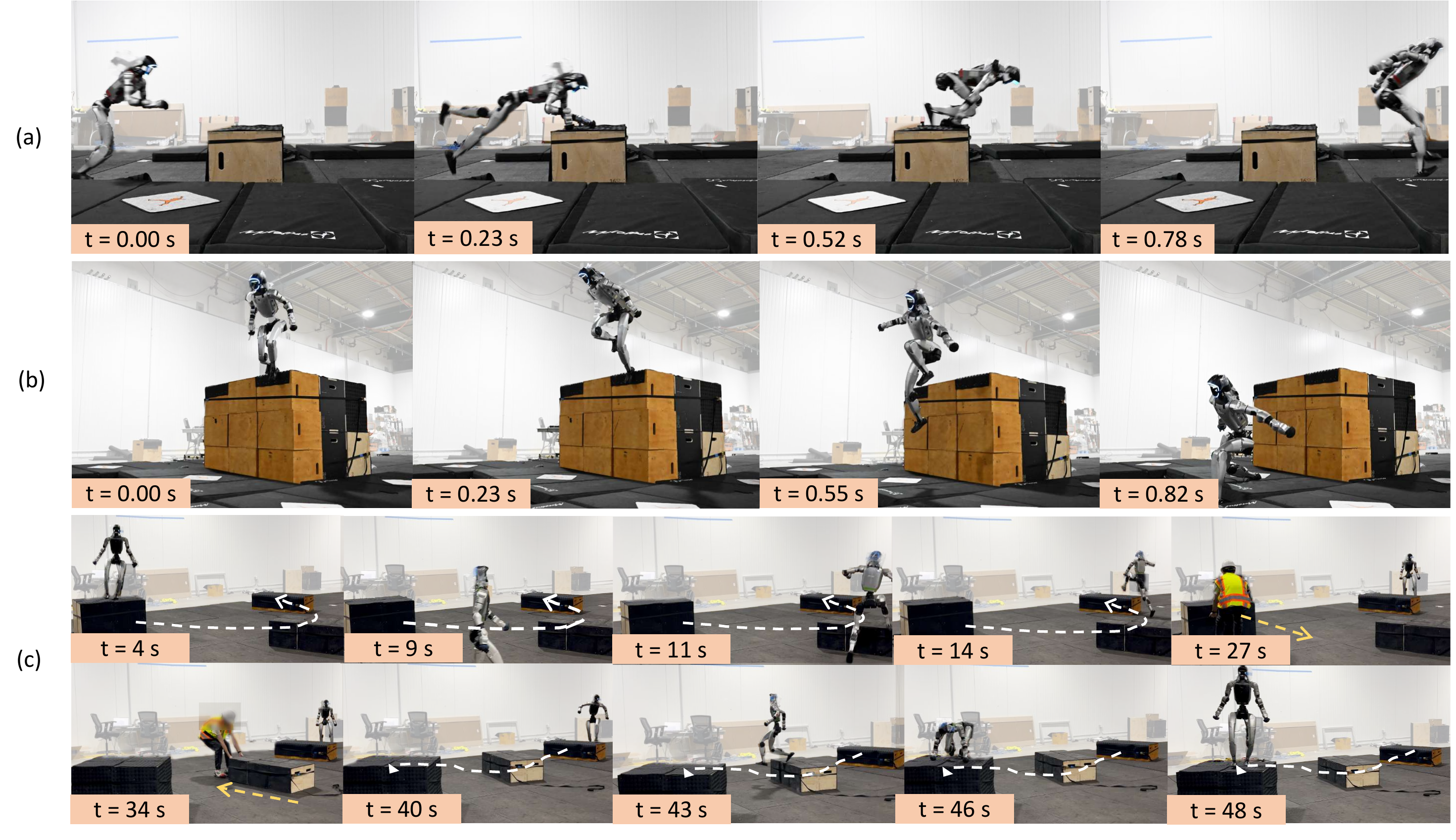}
    \caption{Hardware results demonstrating agile, long-horizon parkour behaviors, including (a) a cat vault, (b) a drop landing from a 1.25\,m wall, and (c) a 48-second terrain traversal with online adaptation to real-time obstacle displacement.}
    \label{fig:hardware_results}
    \vspace{-8pt}
\end{figure*}

We evaluate our system on real-world parkour tasks requiring both highly-dynamic individual skills, long-horizon multi-skill composition and adaptation to environmental changes. All skill execution is autonomous, while only simple 2D velocity commands are provided for navigation.

\subsubsection{Human-Level Agility}
We first demonstrate that the robot can execute highly dynamic parkour skills, including a direct comparison with a human parkourist on a challenging high-wall climb.

\textbf{High-Wall Climb with Human Comparison.}
We compare the robot’s high-wall climb against a human performing the same maneuver~\cite{Salgadopk}.
While often considered a fundamental parkour technique, the high-wall climb demands substantial upper-body strength and precise whole-body coordination, and remains difficult for untrained individuals. 
Despite this, the robot successfully performs the climb at a pace comparable to the human. As shown in Fig.~\ref{fig:climb_134}, the robot executes a fast and coherent sequence with closely matched timing across key events (toe-off $\rightarrow$ pull-up $\rightarrow$ swing $\rightarrow$ stable stand). For a 1.25\,m wall (96\% of the robot’s height), the robot climbs onto the platform in 3.63\,s measured from toe-off.\looseness=-1

\textbf{Additional Parkour Skills.}
We demonstrate additional highly dynamic parkour skills that require rapid contact transitions and momentum preservation.
As shown in Fig.~\ref{fig:hardware_results}(a), 
the robot clears a $0.4$\,m-high, $0.5$\,m-long obstacle within $0.8$\,s from toe-off to toe-on, while covering more than $2$\,m forward ($154\%$ of its height).
The motion reaches a peak forward speed of 3.41\,m/s with an average speed of 2.53\,m/s, highlighting its effective momentum preservation across the contact.
Fig.~\ref{fig:hardware_results}(b) shows a drop landing from a 1.25\,m  platform.
Upon landing, the robot flexes its lower-body joints to absorb impact and stabilize its posture.


\subsubsection{Multi-Obstacle Course}
A key strength of our framework is that the policy generalizes to complex, multi-obstacle courses despite the training data containing only single-obstacle traversal. This capability emerges from motion-matching-based composition, which synthesizes long-horizon reference trajectories that explicitly chain skills through shared locomotion segments and expose the policy to diverse approach distances and timings. As a result, the policy learns to execute skills reliably across varying obstacle sequences without explicit multi-obstacle supervision.

\textbf{Various Skills and Adaptivity to Obstacle Changes.}
As shown in Fig.~\ref{fig:hardware_results}(c), the robot composes multiple skills, including stepping, low and high wall climb, into a continuous run over courses with several obstacles. The visuomotor policy generates transitions online, enabling smooth skill switching throughout the traversal.
We further demonstrate closed-loop adaptivity by randomly displacing multiple obstacles by approximately 0.5\,m during execution. The policy adapts by adjusting its approach and maneuver timing, allowing the robot to continue and complete the traversal in response to these obstacle changes. These results demonstrate the adaptivity of our policy in long-horizon terrain traversal.

\subsection{Quantitative Results in Simulation}
\subsubsection{Experiment Setup}
We evaluate all methods using success rate on parkour traversal tasks.
Each task requires the robot to move forward at a fixed command speed (1.0\,m/s or 2.0\,m/s) and clear a single obstacle of a specified height and $20^\circ$ yaw randomization.
To vary approach conditions, the humanoid is initialized at a random distance in front of the obstacle: for 1\,m/s tasks, distances are sampled uniformly from 1.5\,m to 3.0\,m; for 2\,m/s tasks, from 3.0\,m to 4.5\,m. For each task, we evaluate 100 obstacle instances with different initial distances.
A trial is successful if the robot traverses the obstacle and travels an additional 1.5\,m without falling within a fixed time horizon. We report average success rates of 5 trials per obstacle per task (500 trials total per task). We train all variants with the full skill set, with details in Appx.\ref{appx:distillation}.

\subsubsection{Baseline Comparison}
We evaluate the contribution of three key components: human reference motions, motion-matching-based skill composition, and the two-stage teacher-student training framework, by comparing our method against the following baselines, as shown in \cref{tab:baseline_comparison}.
\begin{itemize}
    \item \textbf{Velocity Tracking.}
    We train a humanoid to traverse terrain using IsaacLab’s  \cite{mittal2025isaac} standard velocity-tracking RL pipeline.
    The policy is learned purely with reward shaping, without any human reference motion. 

    \item \textbf{Uncomposed Motion Data.}
    This baseline removes motion matching, training instead on uncomposed locomotion data and atomic parkour skill clips.

    \item \textbf{End-to-end Depth Policy.}
    This baseline removes distillation and trains a single depth-based visuomotor policy end-to-end on the motion-matching data, using the same observations as the student and the same motion-tracking reward as the experts.
\end{itemize}

\begin{table}[t]
\centering
\caption{Baseline success rate on parkour tasks with different commanded speeds and obstacle heights.}
\label{tab:baseline_comparison}
\setlength{\tabcolsep}{6pt}
\resizebox{\linewidth}{!}{%

\begin{tabular}{lcccccc}
\toprule
Commanded Velocity &
\multicolumn{3}{c}{\textbf{1.0\,m/s}} &
\multicolumn{3}{c}{\textbf{2.0\,m/s}} \\
\cmidrule(lr){2-4} \cmidrule(lr){5-7}
& 36\,cm & 58\,cm & 76\,cm
& 36\,cm & 58\,cm & 76\,cm \\
\midrule
Velocity Tracking             & \textbf{1.00} & 0.00 & 0.00 & \textbf{1.00} & 0.00 & 0.00 \\
Uncomposed Data   & 0.06 & 0.02 & 0.00 & 0.37 & 0.27 & 0.07 \\
End-to-end Depth  & 0.95 & 0.07 & 0.08 & 0.78 & 0.19 & 0.14 \\
\midrule
\textbf{Ours}             & \textbf{1.00} & \textbf{0.99} & \textbf{0.95}
                          & \textbf{1.00} & \textbf{0.99} & \textbf{0.95} \\
\bottomrule
\end{tabular}
}
\vspace{-10pt}
\end{table}

We found that the \textbf{velocity-tracking} baseline achieves similar performance to prior reward-shaping works~\cite{zhuang2024humanoid, long2025learning} and succeeds in traversing the 36 cm obstacle, but fails on higher obstacles. Specifically, it largely relies on foot-only stepping and does not discover whole-body climbing strategies that use the arms for support,
highlighting the limitations of reward-shaping RL alone for highly dynamic parkour. 

The \textbf{uncomposed motion data} baseline performs poorly despite access to atomic skills, showing that isolated motions are insufficient. A common failure mode is that the robot walks up to an obstacle but fails to climb or jump over it. Without explicit long-horizon composition, the policy neither experiences skill transitions during training nor observes obstacles during the walking phase to prepare the appropriate upcoming skill. In comparison, our motion-matching-based approach addresses this limitation by both generating coherent long-horizon skill composition with smooth transitions and exposing the policy to diverse visual contexts during skill execution.

While \textbf{end-to-end depth-based} training can handle low obstacles, its performance degrades on more challenging tasks, suggesting difficulty in RL exploration when training from scratch. In contrast, our expert-distillation pipeline 
achieves substantially higher success rates across obstacle heights, particularly for highly dynamic skills.

\subsubsection{Ablation Study}
We conduct ablation studies to study the effect of motion-matching reference data density, training scalability, and the role of RL during policy distillation (\cref{tab:ablation}). \looseness=-1



\begin{table}[t]
\centering
\caption{Success rate on parkour tasks with different motion matching densities and RL strategies during distillation.}
\label{tab:ablation}
\setlength{\tabcolsep}{6pt}
\resizebox{\linewidth}{!}{%
\begin{tabular}{lcccccc}
\toprule
\textbf{Method} &
\multicolumn{3}{c}{\textbf{1.0\,m/s}} &
\multicolumn{3}{c}{\textbf{2.0\,m/s}} \\
\cmidrule(lr){2-4} \cmidrule(lr){5-7}
& 58\,cm & 76\,cm & 94\,cm 
& 36\,cm & 58\,cm & 76\,cm \\
\midrule
Extreme Distances & 0.99 & 0.62 & 0.64 & 0.98 &  0.60 & 0.58 \\
Half Density & 0.95 & 0.32 & 0.57 & 0.99 & 0.85 & 0.81\\
\midrule
DAgger Only & 0.16 & 0.03 & 0.12 & 0.63 & 0.09 & 0.10 \\
DAgger \& Alive Reward & \textbf{1.00} & 0.90 & 0.96 & 0.94 & 0.91 & 0.84\\
DAgger \& Root Tracking & \textbf{1.00} & 0.79 & 0.75 & \textbf{1.00} &0.92 & 0.87 \\
\midrule
1/4 Training Envs & 0.97 & 0.00 & 0.59 & 0.94 & 0.65 & 0.58\\
1/2 Training Envs & 0.94 & 0.60 & 0.68 & 0.97 & 0.79 & 0.75 \\
3-layer MLP  & 0.99 & 0.02 & 0.00 & 0.98 & 0.89 & 0.81\\
4-layer MLP  & \textbf{1.00} & 0.94 & 0.08 & 1.00 & 0.94 & 0.88\\

\midrule
\textbf{Ours}             & 0.99 & \textbf{0.95} & \textbf{1.00}
                          & \textbf{1.00} & \textbf{0.98} & \textbf{0.90} \\
\bottomrule
\end{tabular}
}
\vspace{-8pt}
\end{table}

\textbf{Motion Matching Density. }
We hypothesize that diversity in motion-matching data, especially approach distances, is critical for accurate timing and task success. To test this, we ablate approach-distance coverage in the reference dataset:

\begin{itemize}
    \item \textbf{Extreme Distances.} Only minimum and maximum approach distances. 
    \item \textbf{Half Density.} Randomly selected half of the full motion-matching data. 
\end{itemize}

Using \textbf{Extreme distances} data leads to reduced success rates across all tasks, as the policy fails to generalize to intermediate distances where contact-timing is critical. Training on \textbf{Half density} data generally yields lower success rates on harder skills, especially when the remaining samples are skewed toward one end of the distance range. For example, in the 1.0\,m/s climbing task on 76\,cm and 94\,cm obstacles, reduced local density leads to unreliable hand-placement timing. In contrast, the full dataset densely covers approach conditions, enabling robust skill execution across varying approach distances. \looseness=-1

\textbf{Training Scalability.} 
We ablate the number of parallel training environments and model capacity to assess the scalability.
\begin{itemize}
\item \textbf{1/4 Training Envs.} Use 1/4 of the training environments. 
\item \textbf{1/2 Training Envs.} Use 1/2 of the training environments. 
\item \textbf{3-Layer MLP.} Use a 3-layer MLP with hidden sizes of [512, 256, 128].
\item \textbf{4-Layer MLP.} Use a 4-layer MLP with hidden sizes of [1024, 512, 256, 128].
\end{itemize}

Unlike training from scratch, where additional rollouts often yield diminishing returns due to exploration limits, our distillation framework scales favorably with both model capacity and rollout throughput. Increasing the number of parallel environments or using a deeper network generally improves success, especially on more challenging parkour tasks.

\textbf{RL in Distillation. }
We ablate the RL objective and its reward design in the distillation stage to understand its role.
\begin{itemize}
\item \textbf{DAgger Only.} Remove the RL loss during distillation.\item \textbf{DAgger + RL Alive Reward.} Use only an alive/progress reward, without motion-tracking terms.
\item \textbf{DAgger + RL Root Tracking Reward.} Use a root-tracking reward instead of full whole-body tracking.
\end{itemize}

We find that RL is critical for effective distillation. The \textbf{DAgger only} student exhibits a clear performance drop, indicating that DAgger alone is insufficient to capture highly dynamic skills even with strong experts. 
For example, on the 76\,cm obstacle, the DAgger student consistently stalls at the pull-up phase: although it learns the accurate hand placement, it fails to produce the brief, high-magnitude torque burst needed to lift the torso. As discussed in Sec.~\ref{sec:distillation_method}, this likely occurs because the decisive torque burst spans only a few timesteps, and per-step imitation loss barely penalizes slightly underestimated actions. In contrast, since RL accounts for episode success, it encourages torque bursts that are more likely to complete the pull-up, yielding both higher reward and lower DAgger loss.

We further evaluate how sensitive the DAgger+RL stage is to reward design.
Interestingly, using \textbf{root tracking} or even only an \textbf{alive reward} achieves success rates comparable to whole-body tracking on difficult skills. This suggests that, when co-trained with DAgger, RL is relatively robust to reward choice and mainly acts as a success-driven exploitation signal that compensates for DAgger’s underestimation, rather than relying on detailed task-specific shaping. Accordingly, while we use whole-body tracking in this work, a simple alive reward may suffice when scaling to larger skill sets.

In addition, our approach differs from prior work that first trains a strong DAgger policy and then applies a separate RL fine-tuning stage~\cite{rudin2025parkour}. Here, we use RL during distillation to correct imitation-induced conservatism and improve skill learning. We also find that the DAgger term must remain active throughout training: if we drop the DAgger loss after the curriculum and continue with pure RL, the policy often develops jittery, unnatural behaviors, suggesting that in a high-dimensional action space the behavior cloning objective provides a critical regularization for RL.






\section{Conclusion} 
We have presented Perceptive Humanoid Parkour, a modular framework that enables humanoid robots to autonomously execute long-horizon, highly dynamic parkour behaviors using onboard perception. By combining motion-matching-based skill composition with a teacher-student RL pipeline, our approach preserves the agility of human motions while enabling perception-driven adaptation to diverse obstacles. We find that dense motion matching is critical in providing coherent long-horizon references and exposes the policy to a wide range of approach conditions, 
while augmenting distillation with RL transfers the capability from single-skill, privileged-information experts to the multi-skill, depth-based student efficiently. 
Through extensive simulation studies and zero-shot deployment on a Unitree G1 robot, we demonstrate state-of-the-art agile, adaptive, whole-body parkour in the real world.


While our pipeline enables long-horizon, highly dynamic
humanoid parkour, it currently lacks semantic scene understanding. 
Incorporating richer conditioning signals, such as language, could enable finer control over diversity and styles. In addition, our real-world capabilities are constrained by perception and hardware. With a short-range, narrow field-of-view camera at a high running speed, obstacle geometries may not be visible sufficiently early, forcing the robot to commit under perceptual ambiguity. Improved sensing and semantic scene understanding could reduce this ambiguity and support richer context reasoning. Finally, our hardware lacks sufficiently strong hands or grippers for interactions with edges and bars to be tested, preventing more extreme climbing beyond the robot’s height or hanging maneuvers.



\bibliographystyle{plainnat}
\bibliography{references}

\clearpage\newpage
\appendix
\subsection{Motion Matching Implementation Details}
\label{appx:mm_implementation}

This section provides implementation details for the motion matching procedure used to synthesize long-horizon parkour reference trajectories.

\subsubsection{Motion Database and Feature Precomputation}
All motion clips are first retargeted to a 29-DOF Unitree G1 humanoid using OmniRetarget~\cite{yang2025omniretarget} and represented as frame sequences.
At each frame $i$, we store the robot configuration $\bm{q}_i = (\bm{p}_i, \bm{r}_i, \bm{\theta}_i)$, consisting of the root translation $\bm{p}_i \in \mathbb{R}^{3}$, root quaternion $\bm{r}_i \in \mathbb{R}^{4}$, and joint angles $\bm{\theta}_i \in \mathbb{R}^{29}$.
For each frame, we also precompute a matching feature vector $\bm{x}_i$ derived from $\bm{q}_i$.
Following~\cite{holden2020mm}, $\bm{x}_i \in \mathbb{R}^{27}$ is expressed in the character’s local coordinate frame and consists of:
\begin{itemize}
    \item \textbf{Future root trajectory $\bm{t}_i \in \mathbb{R}^{12}$:} Planar root positions and facing directions at 0.33s, 0.67s, and 1s into the future.
    \item \textbf{Local foot state $\bm{f}_i \in \mathbb{R}^{12}$:} Positions and linear velocities of the left and right feet expressed in the root frame.
    \item \textbf{Root velocity $\bm{h}_i \in \mathbb{R}^{3}$:} Root linear velocity.
\end{itemize}
To improve data coverage, we augment the motion database by mirroring all motion clips.
For parkour motion clips, we manually fit a box-shaped terrain aligned with each motion.

\subsubsection{Query Feature Construction}
At runtime, a query feature $\hat{\bm{x}}_t$ is constructed from  the current robot configuration $\bm{q}_t$ and a 2D velocity command.
We first extract the kinematic features from $\bm{q}_t$ to form the pose-based part of the query, namely the local foot state $\hat{\bm{f}}_t$ and the root velocity $\hat{\bm{h}}_t$.
We then compute the short-horizon future root trajectory from the 2D velocity command to form the command-based part $\hat{\bm{t}}_t$.

Following~\cite{holden2020mm}, we convert the 2D velocity command into a future root trajectory using a critically damped spring model.
We apply the spring to (i) the 2D root velocity and (ii) the root heading direction.
The target 2D velocity is set to the commanded 2D velocity $\bm{u}^{\text{cmd}}_t \in \mathbb{R}^2$.
The target heading $\psi^{\text{cmd}}_t$ is set to $ \mathrm{atan2}(u^{\text{cmd}}_{t,y},u^{\text{cmd}}_{t,x})$.

\textbf{Critically damped spring closed form.}
Let $\bm{s}$ denote the spring position and $\dot{\bm{s}}$ its velocity, with goal $\bm{s}_{\text{goal}}$ and damping parameter $y>0$.
Define $\bm{j}_0=\bm{s}_0-\bm{s}_{\text{goal}}$ and $\bm{j}_1=\dot{\bm{s}}_0 + y\,\bm{j}_0$.
Then the spring state at any future time $\tau$ admits the closed form
\begin{equation}
\bm{s}(\tau)=e^{-y\tau}\big(\bm{j}_0+\tau \bm{j}_1\big)+\bm{s}_{\text{goal}}.
\label{eq:crit_spring}
\end{equation}

\textbf{Planar position from target velocity.}
For planar root translation, we use the spring in velocity space: the spring ``position'' corresponds to planar velocity (and its derivative to acceleration).
We obtain future root positions by integrating the closed-form velocity:
\begin{equation}
\bm{p}(\tau)
=
\bm{p}_0
-\frac{\bm{j}_1}{y^2}e^{-y\tau}
+\frac{-\bm{j}_0-\tau\bm{j}_1}{y}e^{-y\tau}
+\frac{\bm{j}_1}{y^2}
+\frac{\bm{j}_0}{y}
+\bm{u}^{\text{cmd}}_t\,\tau ,
\end{equation}
applied component-wise in the plane.

\textbf{Heading direction from target heading.}
For rotation, we apply the spring directly to the heading angle $\psi$ toward $\psi^{\text{cmd}}_t$ and evaluate the resulting $\psi(\tau)$ at the same horizons via \cref{eq:crit_spring} (no integration is needed).

We evaluate at $\tau\in\{0.33,0.67,1.0\}$\,s, and transform the resulting future positions $\bm{p}(\tau)$ and facing directions into the character's local coordinate frame to form the future trajectory feature $\hat{\bm{t}}_t$.


\subsubsection{Transition Smoothing via Inertialization}
To ensure smooth transitions when switching the playback index to a newly retrieved frame, we adopt inertialization~\cite{bollo2018inertialization}.
The key idea is to compute an offset between the currently playing motion and the target motion at the transition instant, apply this offset after switching so the output remains continuous, and then gradually decay the offset to zero.
We decay this offset using the same critically damped spring model as in~\cref{eq:crit_spring}, but with the goal set to zero.


\subsection{Skill List and Training Implementation Details }
\label{appx:distillation}
\subsubsection{Skill List}
Our motion library includes locomotion and a set of atomic parkour skills.
Locomotion provides a shared transition manifold and includes standing, walking, and running motions spanning commanded speeds from 0.8 to 3.5\,m/s.
Most parkour skills are instantiated at  1.0\,m/s and 2.0\,m/s.
We additionally include a single 3.0\,m/s cat-vault skill to cover extreme-speed vaulting behaviors.
~\cref{tab:motion_data} summarizes the full skill list and the total duration of motion clips for each category.

\subsubsection{Motion Tracking Details}
Specific reward formulations and domain randomization settings used for expert policy learning from~\cite{liao2025beyondmimic} are summarized in Table~\ref{tab:rewardterms} and Table~\ref{tab:domain_rand} for reference.

\subsubsection{Distillation Details}
During student training, we relax the termination conditions relative to the expert to prevent premature termination of valid but mirrored executions. While this improves PPO stability, the student may visit states that are out-of-distribution for the expert policies, which were trained under the original termination thresholds and may not provide meaningful actions in these regimes. In particular, when the student violates the expert’s original termination condition but remains within the relaxed one, querying the expert would yield unreliable supervision. To avoid introducing incorrect DAgger signals, we disable the DAgger loss at such timesteps and rely solely on the PPO objective.

For depth sensing, beyond the aforementioned camera noise, we also add a random depth offset within $\pm3$\,cm and inject i.i.d.\ Gaussian noise with a standard deviation of 3\,cm into the depth observations during training. The onboard depth camera operates at 30\,Hz.

\subsubsection{Training Hyperparameters}
We include all hyperparameters for two-stage training in Table~\ref{tab:train_hyperparameters} for reference. 

        


\begin{table}[t]
\centering
\small
\setlength{\tabcolsep}{6pt}
\renewcommand{\arraystretch}{1.1}
\begin{tabular}{p{0.45\linewidth} c}
\toprule
\textbf{Skill} & \textbf{Duration (s)} \\
\midrule

\multicolumn{2}{l}{\textbf{Locomotion}} \\
Locomotion & 495.5 \\

\midrule
\multicolumn{2}{l}{\textbf{Parkour skills @ 1.0\,m/s}} \\
Step (36 cm) & 2.2 \\
Climb (58 cm) & 12.1 \\
Climb (76 cm) & 8.8 \\
Climb (94 cm) & 10.3 \\

\midrule
\multicolumn{2}{l}{\textbf{Parkour skills @ 2.0\,m/s}} \\
Step (36 cm) & 1.6 \\
Climb (58 cm) & 6.1 \\
Climb (76 cm) & 4.4 \\
Climb (94 cm) & 5.2 \\
Climb (125 cm) & 5.9 \\
Dash Vault & 5.0 \\
Speed Vault & 3.1 \\

\midrule
\multicolumn{2}{l}{\textbf{Parkour skills @ 3.0\,m/s}} \\
Cat Vault & 1.5 \\

\bottomrule
\end{tabular}
\caption{Motion clips used in our motion library.
}
\label{tab:motion_data}
\end{table}

\subsection{Details for Baselines}
\subsubsection{Velocity Tracking Baseline}
To show the importance of human reference motion in our framework, we include a standard reward-shaping velocity-tracking baseline that learns locomotion purely from handcrafted rewards and a terrain curriculum, without any motion imitation or human reference trajectories.
We follow IsaacLab’s standard rough-terrain velocity-tracking recipe using its Unitree G1 rough-terrain configuration\footnote{Code available at \url{https://github.com/isaac-sim/IsaacLab/blob/main/source/isaaclab_tasks/isaaclab_tasks/manager_based/locomotion/velocity/config/g1/rough_env_cfg.py}.}
, which is widely used for humanoid locomotion and terrain traversal and is largely consistent with state-of-the-art reward-shaping-based setups~\cite{hoeller2024anymal}. In alignment with prior works~\cite{zhuang2024humanoid, hoeller2024anymal}, we also employ a terrain curriculum to ease learning. Specifically, we gradually increase terrain difficulty from 0.3\,m to 1.0\,m over 10 levels (with a 2\,m run-up), providing a smooth progression of tasks that helps the policy bootstrap stable locomotion on easier terrain before tackling harder contact and balance challenges as the terrain becomes progressively harder. The curriculum advances the robot to a harder level once it achieves sufficient success on the current level, and moves it back to an easier level if its performance drops.
Notably, different from our student policy which relies on an onboard depth camera for sensing, this baseline directly receives a local terrain height map (i.e., privileged height observations from simulation), which has been shown highly effective in prior parkour systems~\cite{hoeller2024anymal, rudin2025parkour}.


\subsubsection{AMP Baseline}
Since AMP~\cite{peng2021amp} is a popular algorithm for chaining skills with human reference data, we also implemented an AMP baseline by following the \texttt{MimicKit}\footnote{Code available at \url{https://github.com/xbpeng/MimicKit}.}
 AMP implementation released by the original AMP authors. In our experiments, this baseline can walk stably and track the commanded velocity, but it does not perform well on obstacle traversal: it fails on most tasks, especially the harder ones, which is broadly consistent with prior reports that AMP can be difficult to extend to agile motions~\cite{zhang2025add}. At the same time, we recognize that AMP performance can depend strongly on implementation details and careful tuning~\cite{wang2025physhsi}, and we did not have the bandwidth to fully explore this tuning space. For this reason, we do not include AMP in the formal comparison, and instead leave this result as a note for context.

\begin{table}[t]
\centering
\caption{Reward formulation using Gaussian-shaped tracking scores.}
\label{tab:rewardterms}
\resizebox{\columnwidth}{!}{%
\begin{tabular}{llc}
\toprule
\textbf{Reward Terms} & \textbf{Equation} & \textbf{Weight} \\
\midrule
\multicolumn{3}{l}{\emph{Task (Tracking)}}\\
Body Position
& $\exp\!\Big(
-\big( \tfrac{1}{|\mathcal{B}_{\mathrm{target}}|}
\sum_{b\in\mathcal{B}_{\mathrm{target}}}
\|\mathbf{p}^{\mathrm{des}}_{b}-\mathbf{p}_{b}\|^{2} \big) / 0.3^{2}
\Big)$
& $1.0$\\[2mm]
Body Orientation
& $\exp\!\Big(
-\big( \tfrac{1}{|\mathcal{B}_{\mathrm{target}}|}
\sum_{b\in\mathcal{B}_{\mathrm{target}}}
\|\log(R^{\mathrm{des}}_{b}R_{b}^{\top})\|^{2} \big) / 0.4^{2}
\Big)$
& $1.0$\\[2mm]
Body Linear velocity
& $\exp\!\Big(
-\big( \tfrac{1}{|\mathcal{B}_{\mathrm{target}}|}
\sum_{b\in\mathcal{B}_{\mathrm{target}}}
\|\mathbf{v}^{\mathrm{des}}_{b}-\mathbf{v}_{b}\|^{2} \big) / 1.0^{2}
\Big)$
& $1.0$\\[2mm]
Body Angular velocity
& $\exp\!\Big(
-\big( \tfrac{1}{|\mathcal{B}_{\mathrm{target}}|}
\sum_{b\in\mathcal{B}_{\mathrm{target}}}
\|\boldsymbol{\omega}^{\mathrm{des}}_{b}-\boldsymbol{\omega}_{b}\|^{2} \big) / 3.14^{2}
\Big)$
& $1.0$\\[2mm]
Anchor Position
& $\exp\!\Big(
-\|\mathbf{p}^{\mathrm{des}}_{\text{anchor}}-\mathbf{p}_{\text{anchor}}\|^{2} / 0.3^{2}
\Big)$
& $1.0$\\[2mm]
Anchor Orientation
& $\exp\!\Big(
-\|\log(R^{\mathrm{des}}_{\text{anchor}}R_{\text{anchor}}^{\top})\|^{2} / 0.4^{2}
\Big)$
& $1.0$\\[2mm]
\midrule
\multicolumn{3}{l}{\emph{Regularization}}\\
Action smoothness
& $\|\mathbf{a}_{t}-\mathbf{a}_{t-1}\|^{2}$
& $-0.1$\\
Joint position limit
& $\sum_{j=1}^{N}\big[\max(l_j-\theta_j,0)+\max(\theta_j-u_j,0)\big]$
& $-10.0$\\
Undesired self-contacts
& $\sum_{b\notin\mathcal{B}_{\mathrm{ee}}}
\mathbf{1}\!\left[\|f^{\mathrm{self}}_{b}\|>1\,\text{N}\right]$
& $-0.5$\\
\bottomrule
\end{tabular}}
\end{table}

\begin{table}[t]
\centering
\caption{Domain randomization parameters. ($\mathcal{U}[\cdot]$: uniform distribution)}
\label{tab:domain_rand}
\resizebox{\columnwidth}{!}{%
\begin{tabular}{ll}
\toprule
\textbf{Domain Randomization} & \textbf{Sampling Distribution} \\
\midrule
\multicolumn{2}{l}{\emph{Physical parameters}}\\
Static friction coefficients & $\mu_{\text{static}} \sim \mathcal{U}[0.4,\,1.3]$ \\
Dynamic friction coefficients & $\mu_{\text{dynamic}} \sim \mathcal{U}[0.4,\,1.1]$ \\
Restitution coefficient & $e_{\text{rest}} \sim \mathcal{U}[0,\,0.5]$ \\
Default joint positions (except ankle) [rad] & $\Delta\theta^0_j \sim \mathcal{U}[-0.01,\,0.01]$\\
Default ankle joint positions [rad] & $\Delta\theta^0_j \sim \mathcal{U}[-0.03,\,0.03]$\\
Torso COM offset [m] &
$\Delta x\!\sim\!\mathcal{U}[-0.025,0.025],\
\Delta y,\Delta z\!\sim\!\mathcal{U}[-0.05,0.05]$ \\
\midrule
\multicolumn{2}{l}{\emph{Root velocity perturbations}}\\
Root linear vel [m/s] & $v_x,v_y\!\sim\!\mathcal{U}[-0.1,0.1],\ v_z\!\sim\!\mathcal{U}[-0.05,0.05]$ \\
Push duration [s] & $\Delta t \sim \mathcal{U}[1.0,\,3.0]$ \\
Root angular vel [rad/s] & $\omega_x,\omega_y,\omega_z\!\sim\!\mathcal{U}[-0.1,0.1]$ \\
\bottomrule
\end{tabular}}
\end{table}
\begin{table}[]
  \centering
  \caption{Training hyperparameters.}
  \label{tab:train_hyperparameters}
  \resizebox{\columnwidth}{!}{%

  \begin{tabular}{lcc}
    \hline
    Hyperparameter & Motion Tracking & Distillation \\
    \hline
    \multicolumn{3}{c}{Architecture} \\
    \hline
    Actor / Student MLP hidden dims & [512, 256, 128] & [2048, 1024, 512, 256, 128] \\
    Critic MLP hidden dims & [512, 256, 128] & [512, 256, 128] \\
    Activation function & ELU & ELU \\
    Init noise std & 1.0 & 0.01 \\
    Depth backbone & -- & 3-layer CNN + GAP \\
    Depth input resolution & -- & $58\times87$ \\
    Depth output dim & -- & 32 \\
    \hline
    \multicolumn{3}{c}{Training} \\
    \hline
    Steps per environment & 24 & 24 \\
    Max iterations & 20{,}000 & 20{,}000 \\
    Learning rate & $1\times10^{-3}$ & $3\times10^{-4}$ \\
    Schedule & adaptive & adaptive after 1000 iterations \\
    Clip parameter & 0.2 & 0.2 \\
    Entropy coefficient & 0.005 & 0.001 \\
    Discount factor ($\gamma$) & 0.99 & 0.99 \\
    GAE $\lambda$ & 0.95 & 0.95 \\
    Desired KL & 0.01 & 0.01 \\
    Learning epochs & 5 & 2 \\
    Mini-batches & 4 & 96 \\
    Max grad norm & 1.0 & 1.0 \\
    \hline
    \multicolumn{3}{c}{Distillation-Specific} \\
    \hline
    Curriculum end epoch & -- & 10{,}000 \\
    Distill loss type & -- & mse \\
    DAgger loss coefficient & -- & 10.0 \\
    \hline
  \end{tabular}}
\end{table}


\end{document}